\documentclass[11pt]{article}
\usepackage{subcaption}
\usepackage{amsmath,amsfonts,amssymb}
\usepackage{graphicx}
\usepackage{setspace}
\usepackage{csquotes}
\usepackage{hyperref}

\title{On Round-Off Errors and Gaussian Blur in Superresolution and in Image Registration}

\author{Serap A.~Savari \\ Texas A\&M University, College Station, TX 77843-3128, USA}

\begin{document}

\maketitle

\begin{abstract}
  Superresolution theory and techniques seek to recover signals from samples
  in the presence of blur and noise.  Discrete image registration can be an
  approach to fuse information from different sets of samples of the same
  signal.  Quantization errors in the spatial domain are inherent to digital
  images.  We consider superresolution and discrete image registration for
  one-dimensional spatially-limited piecewise constant functions which are
  subject to blur which is Gaussian or a mixture of Gaussians as well as to
  round-off errors.  We describe a signal-dependent measurement matrix
  which captures both types of effects.  For this setting we show that the
  difficulties in determining the discontinuity points from two sets of
  samples even in the absence of other types of noise.  If the samples are
  also subject to statistical noise, then it is necessary to align and
  segment the data sequences to make the most effective inferences about
  the amplitudes and discontinuity points.  Under some conditions on the blur,
  the noise, and the distance between discontinuity points, we prove that we
  can correctly align and determine the first samples following each
  discontinuity point in two data sequences with an approach based on
  dynamic programming.
\end{abstract}

\section{Introduction}
The measurement of dimensional quantities like length, distance, and size
is critical to the advancement of science, engineering, and technology.
Therefore, in working with images the goal of computational superresolution
from measurements obtained from instruments or sensors with limited resolution
is attractive.  Superresolution has generated a large literature with
contributions from multiple disciplines (see, e.g., \cite{tong}. \cite{candes},
\cite{mri}, \cite{fw} and the references therein).  One common approach to
model part of the measurement degradation is to consider the action of the
instrument as a low-pass filter \cite[\S 5.5]{ip}. \cite{candes}.
The Gaussian low-pass filter is a common blur model \cite[\S 5.6]{ip}.
\cite{fw}.  Moreover, digital images are quantized \cite[\S 2.4]{ip}, so
even in the absence of other corruptions information is lost between
the output of the low-pass filter and the corresponding observation in the
spatial domain.  The question is if this can hinder superresolution;
it is known that noise limits superresolution \cite{candes}.
Some prior works on superresolution address the role of the signal-to-noise
ratio or consider discretizing the locations of discontinuity points on a grid
\cite{candes}, \cite{fw}, but we are not aware of the type of
careful analysis of round-off errors of amplitudes that we present here.
Moreover, while papers on superresolution
often treat one set of samples of an underlying signal, there is also an
interest in working with multiple low quality images of the same signal
\cite{nist}, \cite{cheng}, \cite{tong}.  This latter objective generally
requires image registrataion, which also has an extensive literature
\cite{tong}, \cite[\S 2.5]{ip}.

In this paper we consider superresolution and discrete image registration for
one-dimensional spatially limited piecewise constant functions with a uniform
sampling grid and have possible distortions due to blur which is Gaussian or
a mixture of Gaussians as well as round-off errors.
In \cite{savari} and \cite{reference} we examine a limiting version of the
problem where there is no blur and no noise and show that in the absence of
additional assumptions the subinterval relationships among the discontinuity
points can be viewed in terms of the partitions of a unit hypercube.
In \cite{s3}, we extend \cite{savari} and \cite{reference} and consider
image registration for two noisy segments of observations. One observation
of \cite{s3} is that the problem requires segmentation in addition to
alignment because of possible variations in the number of samples from
each region of the support.  We demonstrate that
 cross-correlation template matching \cite[p.~1061]{ip} can perform poorly
 as an alignment technique and use sparsity to develop methods that correctly
 align and segment the data sequences under some conditions on the noise.

 In the extension we consider here, we will describe a signal-dependent
 measurement matrix which captures important effects and serves as the basis
 of our investigations.  In the presence of round-off errors, as the blur
 increases the structure of the measurement matrix changes in ways influenced
 by the minimum distance between discontinuity points, the magnitudes of the
 discontinuities, and the position of the sampling grid relative to the
 discontinuity points.  After introducing the measurement matrix our main
 focus will be in a regime of small blur where the measurement matrix is more
 complex than its counterpart in the no blur scenario, but it still has a
 comparatively simple structure.

 We will show an example of a spatially limited piecewise constant function
 and two data sequences each consisting of thirteen samples of the function;
 the example has two features.  First, in the case of no statistical noise
 and a knowledge of the amplitudes it illustrates the difficulty in
 determining  all discontinuity points from the two data sequences.  Second,
 in the case of statistical noise it shows that selecting the maximum from
 cross-correlation template matching can perform poorly as an alignment
 technique.  For the case of no blur, we propose in \cite{s3} an alternative
 based on dynamic programming which seeks the most desirable equal length
 subsequences of the two sequences to address alignment and segmentation.
 However, for the blur regime that is the focus of this paper, the problem
 is more challenging because we need to seek clusters of either one element
 or a pair of consecutive elements from each of the data sequences, and we
 focus on the case where the distance between successive discontinuity points
 is at least twice the sampling interval.  Under some conditions on the noise
 and the blur, we prove that the proposed method correctly aligns and
determines the location of the first sample following each discontinuity
 point of the function.

 The plan for the rest of the paper is as follows.  In Section~2, we
 introduce a measurement and a difference matrix.  In the case of no blur
 the sparsity of the difference matrix is the basis for the techniques of
 \cite{s3}.  In this paper, we allow the difference matrix to be more complex,
 and in Section~3 we also discuss the conditions on the blur that satisfy our
 requirements.  In Section~4, we describe the
 example which illustrates how round-off errors can limit the determination of
 discontinuity points and the possible difficulties of applying
   cross-corrleation template matching in the presence of blur and noise.
   In Section~5, we describe our dynamic programming algorithm and prove that
   under some conditions it correctly aligns and determines the first sample
   following each discontinuity point in both data sequences.
  In Section~6, we conclude.

\section{The Measurement and Difference Matrices}
Following \cite{savari}-\cite{s3}, we define our underlying spatially
limited piecewise constant function 
with $m$ regions in its support by 
\begin{displaymath}
g(t) \; = \; \left\{ \begin{array}{ll}
g_j, & D_{j-1}  \leq t < D_j, \; j \in \{1, \ \dots , \ m\} \\
0, & \mbox{otherwise},
\end{array}
\right.
\end{displaymath}
where $D_0 = 0, \ g_1 \neq 0, \; g_m \neq 0,$ and
$g_j \neq g_{j+1}, \ j \in \{1, \ \dots ,
\ m-1\}$. Let $g_0 = g_{m+1} = 0 $.  For this work, suppose that
$g_j$ is an integer multiple of $\frac{1}{256}$ with bounded magnitude
for all $j$.

We assume uniform sampling with a sampling interval of length $T$.
In \cite{savari}-\cite{s3} we assume the distance between successive
discontinuity points exceeds $T$ and that no two discontinuity points have
a distance which is an integer multiple of $T$.

Let
\begin{displaymath}
  h(t) = \frac{e^{-\frac{t^2}{2\sigma^2}}}{\sigma \sqrt{2 \pi}} , \; - \infty
  < t < \infty
\end{displaymath}
represent a pure Gaussian blur, and let 
$\Phi (z) = \int_{- \infty}^{z} \frac{e^{-t^2 / 2}}{\sqrt{2 \pi}} dt.$
If $\tilde{g} (t) = g(t) \ast h(t)$ is a blurred version of $g(t)$ without
round-off errors, then
\begin{eqnarray*}
  \tilde{g} (t) & = & \sum_{j=0}^{m-1} \int_{D_j}^{D_{j+1}} g_{j+1} h(t-u) du \\
  & = & \sum_{j=0}^{m-1} g_{j+1} \left( \Phi \left( \frac{t-D_j}{\sigma}
  \right)
  - \Phi \left( \frac{t-D_{j+1}}{\sigma} \right) \right) \\
  & = & \sum_{j=0}^{m} (g_{j+1}-g_j)
   \Phi \left( \frac{t-D_j}{\sigma} \right),
\end{eqnarray*}
Suppose we observe $N$ samples of $\tilde{g} (t)$ beginning at $t_0 < 0$
and ending at $t_0 +(N-1)T > D_m.$

Let $\tilde{M}$ be an $N \times (m+1)$ ``deformation'' matrix with elements
\begin{displaymath}
  \tilde{M}_{i,j} \; = \; \Phi \left( \frac{t_0 + iT -D_j}{\sigma} \right),
  \; i \in \{0, \ \dots , \ N-1 \},   \; j \in \{0, \ \dots , \ m \}.
\end{displaymath}
Define the {\em difference vector} $g_D$ by
\begin{displaymath}
  g_D \; = \; (g_1-g_0, \ g_2-g_1, \ \dots , \ g_m-g_{m-1}, \ g_{m+1}- g_m)^T.
\end{displaymath}
If we abuse notation and define the vector $\tilde{g}$ by
\begin{displaymath}
  \tilde{g} \; = \; (  \tilde{g} [0], \ \tilde{g} [1], \ \dots , \
  \tilde{g} [N-1])^T
\end{displaymath}
with $\tilde{g} [i] = \tilde{g} (t_0 + iT), \;
  i \in \{0, \ \dots , \ N-1 \}$, then
  \begin{displaymath}
  \tilde{g} \; = \;   \tilde{M} {g}_D .
  \end{displaymath}

  Because of quantization, we cannot hope to see $\tilde{g}$ even in the
  absence of statistical noise.  In the case of no statistical noise
  we instead observe
  \begin{displaymath}
  \gamma \; = \; (  \gamma [0], \ \gamma [1], \ \dots , \
  \gamma [N-1])^T,
  \end{displaymath}
  where we assume $\gamma [i]$ is the closest integer multiple of
$  \frac{1}{256}$ to $\tilde{g} [i]$.  We now have
    \begin{displaymath}
  \gamma \; = \;   {M} {g}_D ;
  \end{displaymath}
    the {\em measurement matrix} $M$ is a corrupted version of $\tilde{M}$.
    $M$ has its simplest forms at extreme values of $\sigma$.  Since we
    assume the magnitude of $g_j$ is bounded for all $j$ and $m+1<N$, it
    is possible to select $\sigma$ so large that
    \begin{displaymath}
      | \Phi \left( \frac{t_0 + iT -D_j}{\sigma} \right) - 0.5 |
      \; < \; \frac{1}{512 (m+1) \max_k |g_k - g_{k-1} |}
    \end{displaymath}
    for all $i \in \{0, \ \dots , \ N-1 \},   \; j \in \{0, \ \dots , \ m \}.$
    Then because of round-off, $M$ is the $N \times (m+1)$ matrix where
    each element is $0.5$ and $\gamma$ is the all-zero vector.  In this
    scenario, it is impossible to recover any information about $g(t)$.

    Next suppose that it is impossible to sample exactly at a discontinuity point
    of $g(t)$.  Then one can select $\sigma$ so small that
\begin{eqnarray*}
    \Phi \left( \frac{t_0 + iT -D_j}{\sigma} \right) & > & 
    1- \frac{1}{512  |g_{j+1} - g_{j} |} \; \mbox{when} \; D_j < t_0+iT \\
    \mbox{and} \;
    \Phi \left( \frac{t_0 + iT -D_j}{\sigma} \right) & < & 
    \frac{1}{512  |g_{j+1} - g_{j} |} \; \mbox{when} \; D_j > t_0+iT 
\end{eqnarray*}
When $\sigma$ is this small, the sample values that form the constraints are
those closest to the discontinuity points because to first order
$\Phi (t) \approx 1 - \frac{e^{-t^2 / 2}}{t \sqrt{2 \pi}} $
when $t$ is large, and when $t$ is negative with large magnitude
$\Phi (t) \approx  \frac{e^{-t^2 / 2}}{|t| \sqrt{2 \pi}} $
\cite[Ch.~26]{handbook}.  In this situation, the measurement matrix is
identical to the one for the case where there is no blur, and it can be
described as follows:
    
Suppose $\eta_0$ samples of $\tilde{g} (t)$ are taken for $t<0$,
$\eta_j$ samples are taken in the region $D_{j-1} < t < D_j, \;
j \in \{1, \ \dots , \ m \}$ and $\eta_{m+1}$ samples are taken for
$t> D_m$.  $\eta_0$ and $\eta_{m+1}$ are only constrained to be at least one,
but the constraints on $\eta_1 , \ \dots , \ \eta_m$ are described in
\cite{savari} and are based on a simple counting argument.
The measurement matrix $M$ consists of $m+2$ blocks of identical rows.
Block 0, i.e., the top block, has $\eta_0$ all-zero rows.
Block $m+1$, i.e., the bottom block, has $\eta_{m+1}$ all-one rows.
Block $k, \; k \in \{1, \ \dots , \ m \}$, has $\eta_{k}$ rows which
begin with $k$ ones and end with $m+1-k$ zeroes.

When $\sigma$ is small enough to result in the preceding measurement matrix,
there will be a family of piecewise constant functions with support
beginning at zero, with the same number of regions in the support, and with
the same amplitudes which result in the same observation $\gamma$.
Therefore, even in the absence of statistical noise superresolution is not
possible in this case, but one can limit the uncertainties about the
locations of the discontinuity points \cite{savari}, \cite{reference}
and describe subinterval relationships when there are multiple data
sequences with samples of the same underlying function.  If $\gamma$ is
corrupted by noise, then for image registration purposes it is more convenient
to work with a {\em difference matrix} $M_D$ because of possible variations
in $\eta_j, \; j \in \{0, \ \dots , \ m+1 \}$.  $M_D$ has the same first row
as $M$ and for $i \in \{1, \ \dots , \ N-1 \}$, row $i$ consists of
row $i$ of $M$ minus row $i-1$ of $M$.  For $j \in \{0, \ 1, \ \dots , \ m \}$,
define
\begin{displaymath}
  \iota (j) \; = \; \sum_{k=0}^j \eta_k.
\end{displaymath}
The elements $M_{D[i,j]}$ of the difference matrix satisfy
\begin{equation}
  M_{D[i,j]} \; = \; \left\{ \begin{array}{ll}
1, & \mbox{if} \; i =   \iota (j) \\
0, & \mbox{otherwise},
\end{array}
\right.
\end{equation}
Observe that
\begin{equation}
  [M_D g_D]_i \; = \; \left\{ \begin{array}{ll}
    g_{j+1}-g_j, & \mbox{if} \; i =   \iota (j) \; \mbox{for some} \;
j \in \{0, \ 1, \ \dots , \ m \}    \\
0, & \mbox{otherwise},
\end{array}
\right.
\end{equation}
Further note that it is straightforward to extend the preceding discussion
to a mixture of Gaussian blurs.  In this case, the deformation matrix
$\tilde{M}$ is the corresponding weighted average of the deformation
matrices associated with the constituent pure Gaussian blurs, and the
measurement matrix $M$ will again be a corrupted version of the deformation
matrix.  Here the mixture of blurs is negligible if the largest blur causes
deformations that are small relative to round-off errors.

Having given an overview of the problem in the case of negligible blur,
it is next of interest to consider a regime where the blur is small but
discernible.

\section{A New Regime}
We have seen that when the blur's effects are negligible in comparison to
round-off errors, every element of the measurement matrix is either zero
or one.  Let us next relax this so that each column of $M$ and each row
of $M_D$ contain at most one element strictly between zero and one;
we call such elements of the measurement matrix {\em critical values}.
We again begin with a discussion of pure Gaussian blur.

For each $j \in \{0, \ 1, \ \dots , \ m \}$ we are interested in the
value $\nu_j$ for which
\begin{equation}
  \Phi \left( \nu_j \right)  \; = \;
    1- \frac{1}{512  |g_{j+1} - g_{j} |}.
\end{equation}
From tables providing the cumulative distribution function of the
standard normal random variable, we see that $\nu_j$ grows slowly with
increasing $|g_{j+1} - g_{j} |$.  For example,
\begin{itemize}
\item if $|g_{j+1} - g_{j} | = 1$, then $2.88 < \nu_j < 2.89$;
\item if $|g_{j+1} - g_{j} | = 2$, then $3.07 < \nu_j < 3.1$;
\item if $|g_{j+1} - g_{j} | = 4$, then $3.26 < \nu_j < 3.3$;
\item if $|g_{j+1} - g_{j} | = 8$, then $3.45 < \nu_j < 3.49$;
\item if $|g_{j+1} - g_{j} | = 16$, then $3.65 < \nu_j < 3.7$;
\item if $|g_{j+1} - g_{j} | = 32$, then $3.8 < \nu_j < 3.85$;
\item if $|g_{j+1} - g_{j} | = 64$, then $4 < \nu_j < 4.05$;
\item if $|g_{j+1} - g_{j} | = 128$, then $4.15 < \nu_j < 4.2$;
\item if $|g_{j+1} - g_{j} | = 256$, then $4.3 < \nu_j < 4.35$;
\item if $|g_{j+1} - g_{j} | = 512$, then $4.45 < \nu_j < 4.5$.
\end{itemize}

For pure Gauusian blur, if
\begin{displaymath}
  \frac{t_0 + iT -D_j}{\sigma} \ > \ \nu_j , \; \mbox{then} \; M_{i,j} = 1,
\end{displaymath}
and if
\begin{displaymath}
  \frac{t_0 + iT -D_j}{\sigma} \ < \ - \nu_j , \; \mbox{then} \; M_{i,j} = 0.
\end{displaymath}
We have the following result.

\noindent {\bf Proposition 1:}
For pure Gaussian blur,  if
\begin{equation}
  \sigma < \frac{0.5 T}{ \max_j \nu_j}
  \end{equation}
    then each column of the measurement matrix has at most one critical value.
    If a weighted average of pure Gaussian blurs satisfies
    $\sigma_k < 0.5 T / \max_j \nu_j$ for all constituent blurs,
    then each column of the measurement matrix has at most one critical value.

    \noindent {\bf Proof:}  Here we discuss the case of pure Gaussian blur.
    The closest two samples to $D_j$ are samples $\iota (j) -1$ and
    $\iota (j)$.  Suppose
    \begin{eqnarray*}
      t_0 + (\iota (j) -1)T - D_j & = & - \alpha_j T \\
            t_0 + \iota (j) T - D_j & = & (1- \alpha_j) T \\
    \end{eqnarray*}
    for some $0 < \alpha_j < 1$.  Then to guarantee that column $j$ has
    at most one critical value, $\sigma$ must be small enough so that at
    most one of $\alpha_j T / \sigma$ and $(1-\alpha_j) T / \sigma$
    is less than $\nu_j$.  Therfore, blur which satisfies (4) ensures that
    each column of $M$ has at most one critical value.

    We will provide the details of the argument for a mixture of Gaussian
    blurs in a longer version of the paper. $\Box$

    The preceding argument also implies the following:

    \noindent {\bf Corollary 2:} 
Given a pure Gaussian blur with $\sigma < 0.5 T / \max_j \nu_j$
or a mixture of Gaussian blurs with $\sigma_k < 0.5 T / \max_j \nu_j$ for all
$k$.  There are three possible forms for column $j$ of the measurement matrix:
\begin{itemize}
\item $M_{i,j} = 0$ for $i \leq \iota (j)-1$, and
  $M_{i,j} = 1$ for $i \geq \iota (j)$
\item $M_{i,j} = 0$ for $i \leq \iota (j)-1$, $0.5 < M_{\iota (j),j} < 1$,
  and   $M_{i,j} = 1$ for $i \geq \iota (j)+1$
  \item $M_{i,j} = 0$ for $i \leq \iota (j)-2$, $0 < M_{\iota (j)-1,j} < 0.5$,
  and   $M_{i,j} = 1$ for $i \geq \iota (j)$.
\end{itemize}
Hence, the three possible forms for column $j$ of the difference matrix are:
  \begin{itemize}
  \item $M_{D[\iota (j),j]} = 1, \; M_{D[i,j]} = 0$ for $i \neq \iota (j)$
  \item $M_{D[\iota (j),j]} = M_{\iota (j), j} \in (0.5, \ 1), \;
    M_{D[\iota (j)+1,j]} = 1-M_{\iota (j), j}$, \\ and
    $M_{D[i,j]} = 0$ for $i \notin \{\iota (j), \ \iota (j)+1 \}.$
      \item $M_{D[\iota (j)-1,j]} = M_{\iota (j)-1, j} \in (0, \ 0.5), \;
    M_{D[\iota (j),j]} = 1-M_{\iota (j)-1, j}$, \\ and
    $M_{D[i,j]} = 0$ for $i \notin \{\iota (j) -1, \ \iota (j) \}.$
  \end{itemize}
  The nature of the rows of the measurement and difference matrices is
  also essential for the analysis.  In a longer version of the paper we
  provide the proof for the following result.

  \noindent {\bf Proposition 3:}
Given a pure Gaussian blur with $\sigma < 0.5 T / \max_j \nu_j$
or a mixture of Gaussian blurs with $\sigma_k < 0.5 T / \max_j \nu_j$ for all
$k$.  Then no row of the measurement matrix contains more than one component
which is a critical value.  Furthermore, if the minimum distance between
discontinuity points exceeds $2T$, then any row of $M_D$ has at most one
nonzero component.

In \cite{s3} we work with $M_D g_D$ and use (2).  In this regime of
larger blur, the $j^{\mbox{th}}$ component of $g_D$ can either appear
in one element of $M_D g_D$ or be divided between two consecutive components
of $M_D g_D$.  Therefore, we wish to better understand
$M_D g_D$ and particularly cases with less sparsity within $M_D g_D$.
In a longer version of the paper, we provide the proof of the following result.

  \noindent {\bf Theorem 4:}
Given a pure Gaussian blur with $\sigma < 0.5 T / \max_j \nu_j$
or a mixture of Gaussian blurs with $\sigma_k < 0.5 T / \max_j \nu_j$ for all
$k$.  If the minimum distance between discontinuity points exceeds $2T$,
then
\begin{displaymath}
  [M_D g_D]_i \; = \; \left\{ \begin{array}{ll}
 M_{i,j}  ( g_{j+1}-g_j), & \mbox{if for some} \;
 j \in \{0, \ 1, \ \dots , \ m \}  \\
 \mbox{} & \mbox{either}  \; i =   \iota (j) -1 \\
\mbox{} & \mbox{or both} \; i =   \iota (j) \; \mbox{and} \; M_{i-1,j} = 0
\\
(1- M_{i,j})  ( g_{j+1}-g_j), & \mbox{if for some} \;
j \in \{0, \ 1, \ \dots , \ m \}  \\
\mbox{} & \mbox{either}  \; i =   \iota (j) +1 \\
\mbox{} & \mbox{or both} \; i =   \iota (j) \; \mbox{and} \; M_{i-1,j} > 0
 \\
0, & \mbox{otherwise}.
\end{array}
\right.
  \end{displaymath}
  Regarding sparsity,
  \begin{itemize}
  \item if $M_{D[\iota (j), j]} = 1$ for some $j$, then it is impossible
    for both $[M_D g_D]_{\iota (j)-1}$ and $[M_D g_D]_{\iota (j)+1}$
    to be nonzero.  Moreover, if $[M_D g_D]_{\iota (j)-1} \neq 0,$
    then $D_j - D_{j-1} <2.5T$
    and $|[M_D g_D]_{\iota (j)-2} + [M_D g_D]_{\iota (j)-1}| >
    |[M_D g_D]_{\iota (j)} |$; if $[M_D g_D]_{\iota (j)+1} \neq 0,$
    then $D_{j+1} - D_{j} <2.5T$
    and $|[M_D g_D]_{\iota (j)+1} + [M_D g_D]_{\iota (j)+2}| >
    |[M_D g_D]_{\iota (j)}|.$
  \item if for some $j, \; \iota (j+1) = \iota (j)+2$ and either both
    $M_{\iota (j)-1, j} >0$ and     $M_{\iota (j)+1, j+1} >0$
    or both $M_{\iota (j), j} <1$ and     $M_{\iota (j)+2, j+1} <1$,
    then  $D_{j+1} - D_{j} <2.5T$.
      \end{itemize}
  
    \section{A Small Example}
  Consider the underlying spatially-limited piecewise constant function
  \begin{displaymath}
g(t) \; = \; \left\{ \begin{array}{ll}
1, & 0 \leq t < 2.44T, \; 5.01T \leq t < 7.42T \\
-1, & 2.44T \leq t < 5.01T, \; 7.42T \leq t < 9.43T \\
0, & \mbox{otherwise.}
\end{array}
\right.
  \end{displaymath}
  Suppose we have pure Gaussian blur with $\sigma = 0.125T$,
  the first sequence $\gamma_1$   of thirteen noiseless samples of
  $\tilde{g}(t)$ has its first sample at $t=-0.98T$,   and the second sequence
  $\gamma_2$   of thirteen noiseless samples of $\tilde{g}(t)$ has its first
  sample at $t=-0.4T$.
  Then $\gamma_1$ and $\gamma_2$ are
  \begin{eqnarray*}
    \gamma_1 & = & (\gamma_1 [0], \dots ,  \gamma_1 [12]) \\
    &= &  \left(0, \ \frac{144}{256}, \ 1, \ 1, \ -1, \ -1, \ \frac{16}{256},
    \ 1, \ 1, \ -1, \ -1, \ 0, \ 0 \right) \\
    \gamma_2 & = & (\gamma_2 [0], \dots ,  \gamma_2 [12]) \\
    &= & \left(0, \ 1, \ 1, \ -\frac{205}{256}, \ \ -1, \ -1, \ 1, \ 1,
    -\frac{218}{256}, \ -1, \ - \frac{22}{256}, \ 0, \ 0 \right).
  \end{eqnarray*}

  Suppose that we know the underlying function has four regions and we
  know their amplitudes.  Also, first suppose that we believe that there is
  a pure Gaussian blur, but we do not know the value of $\sigma$.
  What can we infer?

  Let us begin by considering $\gamma_1$ by itself.
  \begin{itemize}
  \item Since $\gamma_1 [0] = 0$  it folllows that
    $\Phi \left( \frac{t_1}{\sigma} \right) < \frac{1}{512}$ and
    $t_1 < -2.88 \sigma .$
  \item Since $\gamma_1 [1] = \frac{144}{256},$ it follows that
    $\frac{143.5}{256} < \Phi \left( \frac{t_1 +T}{\sigma} \right) <
    \frac{144.5}{256}$.  Based on a standard normal table, we have that
    $0.15 \sigma < t_1+T < 0.17 \sigma$.
  \end{itemize}
  In a similar manner we proceed through the rest of $\gamma_1$.  By using
  our knowledge about $t_1$ and $t_1 +T$, we obtain
  \begin{eqnarray*}
    2T + 3.22 \sigma & < & D_1 \; < \; 3T-2.9 \sigma \\
    5T + 0.06 \sigma & < & D_2 \; < \; 5T+0.1 \sigma \\
    7T + 3.22 \sigma & < & D_3 \; < \; 8T-2.9 \sigma \\
    \mbox{and} \;     9T + 3.03 \sigma & < & D_4 \; < \; 10T-2.71 \sigma .
  \end{eqnarray*}

  Next, let us consider $\gamma_2$ by itself.  Then we can infer
  \begin{eqnarray*}
    -T+2.88 \sigma & < & t_2 \; < \; - 2.88 \sigma \\
    2T + 1.59 \sigma & < & D_1 \; < \; 3T-4.15 \sigma \\
    4T + 5.95 \sigma & < & D_2 \; < \; 6T-5.95 \sigma \\
    7T + 1.42 \sigma & < & D_3 \; < \; 8T-4.31 \sigma \\
    \mbox{and} \;     9T + 1.5 \sigma & < & D_4 \; < \; 10T-4.23 \sigma .
  \end{eqnarray*}
  In combining
  information from $\gamma_1$ and $\gamma_2$, we can improve our estimates.
  For example, we can infer
  \begin{eqnarray*}
    -T+4.49 \sigma & < & t_2 \; < \; - 2.97 \sigma \\
    2T + 3.22 \sigma & < & D_1 \; < \; 3T-4.24 \sigma \\
    5T + 0.06 \sigma & < & D_2 \; < \; 5T+0.1 \sigma \\
    7T + 3.22 \sigma & < & D_3 \; < \; 8T-4.4 \sigma \\
    9T + 3.11 \sigma & < & D_4 \; < \; 10T-4.32 \sigma 
  \end{eqnarray*}
and $\sigma < \frac{T}{7.62}$.  However, there are additional
  difficulties even in the absence of statistical noise.  If we enlarge our
  belief about possible blur models to a mixture of Gaussian blurs, then
  there will be increased uncertainty about the locations of discontinuity
  points.  For example, a mixture of $\frac{1}{64}$ no blur with
  $\frac{63}{64}$ pure Gaussian blur with unknown $\sigma$ will slightly
  change the inferences.  

  Next suppose $e_k  = ( e_k [0], \ \dots , \ e_k [12] ), \; k \in \{1, \ 2\}$,
  denotes additive noise corrupting $\gamma_k$.
Let the underlying noise model for some $x \in [0, \ 0.5]$ which is an
integer multiple of $\frac{1}{256}$ be
  \begin{displaymath}
    P[e_k [i] = x] \; = \; P[e_k [i] = -x] \; = \; 0.5, \; k \in \{ 1, \ 2\},
    i \in \{ 0, \ 1, \ \dots , \ 12 \}.
  \end{displaymath}

  We consider the following error vectors:
    \begin{eqnarray*}
      e_1 & = &
      (x, \ x, \ -x, \ -x, \ x, \ x, \ x, \ -x, \ -x, \ -x, \ -x, \ x, \ x) \\
      e_2 & = &
   (-x, \ -x, \ -x, \ x, \ -x, \ -x, \ -x, \ -x, \ x, \ x,\ -x, \ -x, \ -x) .
    \end{eqnarray*}

    Our observed sequences of corrupted samples
    $y_k  = ( y_k [0], \ \dots , \ y_k [12] ), \; k \in \{1, \ 2\}$ satisfy
    \begin{displaymath}
      y_k [i] \; = \; \gamma_k [i] + e_k [i] ,  \; k \in \{ 1, \ 2\},
    i \in \{ 0, \ 1,  \ \dots , \ 12 \};
    \end{displaymath}
    assume $y_1 [i] = y_2 [i] = 0, \; i \not\in \{ 0, \ 1, \ \dots , \ 12 \}$.
    To consider cross-correlation template matching, let
    \begin{displaymath}
      r_{y,12} [k] \; = \; \sum_i y_1 [i] y_2 [i+k] , \; k \in
      \{-12, \ -11, \ \dots , \ 11, \ 12\}.
    \end{displaymath}

    For $0 \leq x \leq 0.5,$
    \begin{displaymath}
      \arg \max_k  r_{y,12} [k]
      \; = \; \left\{ \begin{array}{ll}
        -1, & 0 \leq x < \frac{682}{2483}
        \\
        -5, & \frac{682}{2483} < x \leq 0.5.
\end{array}
\right.
\end{displaymath}
    At $x=0$, cross-correlation template matching recommends that we shift
    $\gamma_2$ by one to the right relative to $\gamma_1$, which is incorrect.
  At $x=\frac{71}{256}$, the cross-correlation template matching scheme
 recommends that we shift
    $\gamma_2$ by five to the right relative to $\gamma_1$, which would
lead to    the loss of a considerable portion of the support of $g(t)$.
    Moreover, $\gamma_1$ and $\gamma_2$ imply that $y_1$ and $y_2$
    need segmentation.  Therefore, we seek an alternative.
In what follows we will use the following notation.
Define the difference sequences
$d_k = ( d_k [0], \ \dots , \ d_k [N-1]), \ k \in \{1, \ 2 \}$, by
\begin{displaymath}
d_k [i] \; = \; y_k [i] - y_k [i-1], \; i \in \{0, \ \dots , \ N-1 \}.
\end{displaymath}
To differentiate between the ground truth matrices and vectors for the
two data sequences, we will use the superscript $(k)$.

  \section{A Dynamic Programming Scheme}
  In \cite{s3} we propose techniques to align and segment two noisy sets
  of samples without blur of the same piecewise constant signal and work
  with difference sequences.  Equation~(2) implies that in the case of no
  blur we seek equal length subsequences of the two data sequences under
  certain constraints which are each ideally noisy versions of
  $  g_1, \ g_2-g_1, \ \dots g_m-g_{m-1}, \ - g_m .$  In the case we study here,
  Theorem~4 implies that for each $j$ a noisy version of $g_j - g_{j-1}$
  either resides in a single element of our difference vector or is divided
  into two consecutive elements of our difference vector.
  Moreover, within such a pair it is possible for the sample immediately
  following a discontinuity point to be either in the first or second
  component of the pair.  Because of the possible variations inherent in
  making matches between two difference sequences $d_1$ and $d_2$, it is
  natural to consider a variation of the dynamic programming scheme proposed
  in \cite{s3}.    We formulate a longest path problem from a starting
  vertex to a termination vertex in a directed, acyclic graph to seek the
  most desirable sets of clusters within $d_1$ and $d_2$ subject to both the
  types of constraints needed in \cite{s3} and the constraints of Theorem~4.
  As in the dynamic programming scheme of \cite{s3} we design the graph
  to contain alignment vertices which are neighbors of the starting vertex.
  However, we now work with segmentation pairs of clusters of either a single
  element or two consecutive elements.  We assume $\eta_0 \geq 1$
  so that $d_1 [0]$ and $d_2 [0]$ precede the first samples of the support
  of $\tilde{g} (t)$.
  The alignment vertices are labeled by ordered pairs
  \begin{displaymath}
    (i_1 , \ i_2) \in \{ (0, \ 0), \ (0, \ 1), \ \dots , \ (0, \ N-2), \
    (1, \ 0), \ \dots , (N-2, \ 0) \},
  \end{displaymath}
  where vertex $(i_1 , \ i_2)$ means that we begin examining $d_k, \
  k \in \{1, \ 2\}$, in position $i_k$. We consider here an adaptation of the first edge weight
  function of \cite{s3}; any
  edge from the starting vertex to an alignment vertex has weight zero.
  Each alignment vertex is also neighbors with a subset of the segmentation
  vertices.

  For  the segmentation vertices and edges associated with them we use the
  results from \cite{savari}-\cite{s3} that the number $\eta_j, \
  j \in \{1, \ \dots , \ m \}$, of samples from region $j$ of the support of
  $g(t)$ can differ in two observations by at most one, and the total
  number of samples from the first $j$ regions of the support of $g(t)$ can
  also differ by at most one.  A segmentation pair $(d_1 (i_1 ), \
  d_2 (i_2 )), \; i_1, \ i_2 \in \{1, \ \dots , \ N-1 \}$, marking the
  beginning of the same region in the support of $g(t)$ or in the final
  all-zero region is associated with 48 vertices in the graph labeled
  $(i_1, \ \lambda_1 , \ i_2, \ \lambda_2 , \ n),$
  where $\lambda_1,  \ \lambda_2  \in \{ F, \ S, \ A_b , \ A_e \}$ and
  $n \in \{ 0, \ 1 , \ 2\}$.
  The pair $  (i_k, \ F), \ k \in \{ 1 , \ 2\}$, represents the case in
  $d_k$ where $i_k = \iota^{(k)} (j)$ for some $j$
  and $0.5 < M^{(k)}_{i_k , j} < 1$ so that
  $[M_D^{(k)} g_D]_{i_k +1} \neq 0.$
    The pair $  (i_k, \ S), \ k \in \{ 1 , \ 2\}$, signifies a situation in
  $d_k$ where $i_k = \iota^{(k)} (j)$ for some $j$
  and $0 < M^{(k)}_{i_k -1, j} < 0.5$ so that
  $[M_D^{(k)} g_D]_{i_k -1} \neq 0.$
  The pairs $  (i_k, \ A_b)$
  and $  (i_k, \ A_e), \ k \in \{ 1 , \ 2\}$, respectively denote the
  scenarios where $i_k = \iota^{(k)} (j)$ for some $j$, 
  $[M_D^{(k)} ]_{i_k , j} =1$ and either
  $[M_D^{(k)} g_D]_{i_k -1} = 0$ or
  $[M_D^{(k)} g_D]_{i_k +1} = 0.$
  Within $d_k, \ k \in \{ 1 , \ 2\}$, there are certain transitions from
  $(i_k, \ \lambda_k)$ to   $(i_k^{'}, \ \lambda_k^{'})$ that are
  {\em admissible} based on the minimum distance between discontinuity
  points and Theorem~4:
  \begin{itemize}
  \item We can travel from $(i_k, \ F)$
    to $(i_k +l, \ F)$ for $2 \leq l \leq N-2-i_k$,
    to $(i_k +l, \ S)$ for $3 \leq l \leq N-1-i_k$,
    to $(i_k +l, \ A_b)$ for $3 \leq l \leq N-1-i_k$,
    and to $(i_k +2, \ A_e).$

     \item We can travel from $(i_k, \ S)$ or from $(i_k, \ A_b)$
    to $(i_k +l, \ F)$ for $2 \leq l \leq N-2-i_k$,
    to $(i_k +l, \ S)$ for $2 \leq l \leq N-1-i_k$,
    and    to $(i_k +l, \ A_b)$ for $2 \leq l \leq N-1-i_k$.

         \item We can travel from $(i_k, \ A_e)$
    to $(i_k +l, \ F)$ for $2 \leq l \leq N-2-i_k$,
    to $(i_k +l, \ S)$ for $3 \leq l \leq N-1-i_k$,
and    to $(i_k +l, \ A_b)$ for $2 \leq l \leq N-1-i_k$.
      \end{itemize}
  The vertex $(i_1, \ \lambda_1 , \ i_2, \ \lambda_2 , \ 0),$
  $\lambda_1,  \ \lambda_2  \in \{ F, \ S, \ A_b , \ A_e \}$ 
  can appear on a path from alignment vertex
  $(k_1, \ k_2)$ to the termination vertex if $i_1 - k_1 = i_2 - k_2$,
  the vertex $(i_1, \ \lambda_1 , \ i_2, \ \lambda_2 , \ 1),$
  $\lambda_1,  \ \lambda_2  \in \{ F, \ S, \ A_b , \ A_e \}$ 
  can appear if $i_1 - k_1 = i_2 - k_2 + 1$, 
  and the vertex
  $(i_1, \ \lambda_1 , \ i_2, \ \lambda_2 , \ 2),$
  $\lambda_1,  \ \lambda_2  \in \{ F, \ S, \ A_b , \ A_e \}$
    can appear if $i_1 - k_1 +1 = i_2 - k_2 $.
  If $i_1 = 1$ and/or $i_2 = 1$, then we associate the segmentation pair
  $(d_1 (i_1 ), \   d_2 (i_2 ))$ with the vertices
  $(i_1, \ \lambda_1 , \ i_2, \ \lambda_2 , \ 0),$
  $\lambda_1,  \ \lambda_2  \in \{ F, \ S, \ A_b , \ A_e \}$.
  
  Let $W(i_1, \ \lambda_1 , \ i_2, \ \lambda_2 , \ v)$
  denote the weight of any incoming edge to
  $(i_1, \ \lambda_1 , \ i_2, \ \lambda_2 , \ n),$
  where $\lambda_1,  \ \lambda_2  \in \{ F, \ S, \ A_b , \ A_e \}$ and
  $n \in \{ 0, \ 1 , \ 2\}; \; v>0$
  is a parameter that discourages choosing edges with small magnitudes in an
  optimal path.

  The outgoing edges from alignment vertex $(i_1, \ i_2)$ are to
  the union of the vertices contained in
  \begin{itemize}
\item  $\{(i_1 +l, \ F, \ i_2+l , \ F, \ 0): \; 1 \leq l \leq \min \{N-2-i_1 , \
  N-2-i_2 \} \}$
\item $\{(i_1 +l, \ F, \ i_2+l , \ \lambda_2 , \ 0): \; \lambda_2 \in
  \{ S, \  A_b \}, \ 1 \leq l \leq \min \{N-2-i_1 , \
  N-1-i_2 \} \}$,
\item $\{(i_1 +l, \ \lambda_1 , \ i_2+l , \ F , \ 0): \; \lambda_1 \in
  \{ S, \  A_b \}, \ 1 \leq l \leq \min \{N-1-i_1 , \
  N-2-i_2 \} \}$, and
 \item $\{(i_1 +l, \ \lambda_1 , \ i_2+l , \ \lambda_2 , \ 0): \; \lambda_1, \
  \lambda_2 \in  \{ S, \  A_b \}, \ 1 \leq l \leq \min \{N-1-i_1 , \
  N-1-i_2 \} \}$.
  \end{itemize}

  The outgoing edges from vertex
  $(i_1, \ \lambda_1 , \ i_2, \ \lambda_2 , \ 0), \;
  \lambda_1,  \ \lambda_2  \in \{ F, \ S, \ A_b , \ A_e \}$ 
  are to
  \begin{itemize}
  \item the termination vertex with edge weight zero
  \item   $\{(i_1 +l, \ \lambda_1^{'}, \ i_2+l, \ \lambda_2^{'}, \ 0):
    \lambda_1^{'},  \ \lambda_2^{'} \in \{ F, \ S, \ A_b , \ A_e \}, \ l$
    for which the transitions from $(i_1 , \ \lambda_1)$ to
    $(i_1 +l, \ \lambda_1^{'})$ and from $(i_2 , \ \lambda_2)$ to
    $(i_2+l, \ \lambda_2^{'})$ are both admissible$\}$
   \item   $\{(i_1 +l+1, \ \lambda_1^{'}, \ i_2+l, \ \lambda_2^{'}, \ 1):
    \lambda_1^{'},  \ \lambda_2^{'} \in \{ F, \ S, \ A_b , \ A_e \}, \ l$
    for which the transitions from $(i_1 , \ \lambda_1)$ to
    $(i_1 +l+1, \ \lambda_1^{'})$ and from $(i_2 , \ \lambda_2)$ to
    $(i_2+l, \ \lambda_2^{'})$ are both admissible$\}$
   \item   $\{(i_1 +l, \ \lambda_1^{'}, \ i_2+l+1, \ \lambda_2^{'}, \ 2):
    \lambda_1^{'},  \ \lambda_2^{'} \in \{ F, \ S, \ A_b , \ A_e \}, \ l$
    for which the transitions from $(i_1 , \ \lambda_1)$ to
    $(i_1 +l, \ \lambda_1^{'})$ and from $(i_2 , \ \lambda_2)$ to
    $(i_2+l+1, \ \lambda_2^{'})$ are both admissible$\}$.
    \end{itemize}

  The outgoing edges from vertex
  $(i_1, \ \lambda_1 , \ i_2, \ \lambda_2 , \ 1), \;
  \lambda_1,  \ \lambda_2  \in \{ F, \ S, \ A_b , \ A_e \}$ 
  are to
  \begin{itemize}
  \item the termination vertex with edge weight zero
  \item   $\{(i_1 +l, \ \lambda_1^{'}, \ i_2+l, \ \lambda_2^{'}, \ 1):
    \lambda_1^{'},  \ \lambda_2^{'} \in \{ F, \ S, \ A_b , \ A_e \}, \ l$
    for which the transitions from $(i_1 , \ \lambda_1)$ to
    $(i_1 +l, \ \lambda_1^{'})$ and from $(i_2 , \ \lambda_2)$ to
    $(i_2+l, \ \lambda_2^{'})$ are both admissible$\}$
   \item   $\{(i_1 +l, \ \lambda_1^{'}, \ i_2+l+1, \ \lambda_2^{'}, \ 0):
    \lambda_1^{'},  \ \lambda_2^{'} \in \{ F, \ S, \ A_b , \ A_e \}, \ l$
    for which the transitions from $(i_1 , \ \lambda_1)$ to
    $(i_1 +l, \ \lambda_1^{'})$ and from $(i_2 , \ \lambda_2)$ to
    $(i_2+l+1, \ \lambda_2^{'})$ are both admissible$\}.$
  \end{itemize}

  The outgoing edges from vertex
  $(i_1, \ \lambda_1 , \ i_2, \ \lambda_2 , \ 2), \;
  \lambda_1,  \ \lambda_2  \in \{ F, \ S, \ A_b , \ A_e \}$ 
  are to
  \begin{itemize}
  \item the termination vertex with edge weight zero
  \item   $\{(i_1 +l, \ \lambda_1^{'}, \ i_2+l, \ \lambda_2^{'}, \ 2):
    \lambda_1^{'},  \ \lambda_2^{'} \in \{ F, \ S, \ A_b , \ A_e \}, \ l$
    for which the transitions from $(i_1 , \ \lambda_1)$ to
    $(i_1 +l, \ \lambda_1^{'})$ and from $(i_2 , \ \lambda_2)$ to
    $(i_2+l, \ \lambda_2^{'})$ are both admissible$\}$
   \item   $\{(i_1 +l+1, \ \lambda_1^{'}, \ i_2+l, \ \lambda_2^{'}, \ 0):
    \lambda_1^{'},  \ \lambda_2^{'} \in \{ F, \ S, \ A_b , \ A_e \}, \ l$
    for which the transitions from $(i_1 , \ \lambda_1)$ to
    $(i_1 +l+1, \ \lambda_1^{'})$ and from $(i_2 , \ \lambda_2)$ to
    $(i_2+l, \ \lambda_2^{'})$ are both admissible$\}$.
    \end{itemize}

  We next define an edge weight function.
  For $k \in \{1, \ 2\}$ and $v>0$, let
          \begin{displaymath}
  \omega_k (i, \ F, \ v)          
            = \left\{ \begin{array}{ll}
              d_k [i]+d_k [i+1],        & \mbox{if} \;
              d_k [i] d_k [i+1] > 0, \; |d_k [i]| > |d_k [i+1]| \geq v \\
  0 ,     & \mbox{otherwise,}
\end{array}
\right.
          \end{displaymath}
          \begin{displaymath}
  \omega_k (i, \ S, \ v)          
            = \left\{ \begin{array}{ll}
              d_k [i-1]+d_k [i],        & \mbox{if} \;
              d_k [i-1] d_k [i] > 0, \; |d_k [i]| > |d_k [i-1]| \geq v \\
  0 ,     & \mbox{otherwise,}
\end{array}
\right.
          \end{displaymath}

                    \begin{displaymath}
  \omega_k (i, \ A_b, \ v)          
            = \left\{ \begin{array}{ll}
              d_k [i],        & \mbox{if} \; |d_k [i-1]|<v, \; |d_k [i]| \geq v,
              \\
              \mbox{} & \mbox{and if} \; |d_k [i]| < |d_k [i+1] + d_k [i+2]|
              \; \mbox{ whenever} \; |d_k [i+1]| \geq v \\
  0 ,     & \mbox{otherwise,}
\end{array}
\right.
          \end{displaymath}          
                    \begin{displaymath}
  \omega_k (i, \ A_e, \ v)          
            = \left\{ \begin{array}{ll}
              d_k [i],        & \mbox{if} \; |d_k [i+1]|<v, \; v \leq
              d_k [i]| < |d_k [i-2] + d_k [i-1]| \\
  0 ,     & \mbox{otherwise,}
\end{array}
\right.
          \end{displaymath}          

For $\lambda_1,  \ \lambda_2 \in \{ F, \ S, \ A_b , \ A_e \},$ define
                    \begin{displaymath}
W (i_1, \ \lambda_1, i_2, \ \lambda_2, \ v)          
            = \left\{ \begin{array}{ll}
              1,        & \mbox{if} \; \omega_1 (i_1 , \ \lambda_1 , \  v ) \omega_2 (i_2 , \ \lambda_2 , \  v ) > 0 \\
  0 ,     & \mbox{otherwise,}
\end{array}
\right.
    \end{displaymath}

                    In a longer version of the paper we provide a proof of the the following
                    result.

\noindent {\bf Theorem 5:}
Given a pure Gaussian blur with $\sigma < 0.5 T / \max_j \nu_j$
or a mixture of Gaussian blurs with $\sigma_k < 0.5 T / \max_j \nu_j$ for all
$k$.  Suppose the minimum distance between discontinuity points exceeds $2T$.
Given two difference sequences $d_k, \ k \in \{1, \ 2\}$, with ground truth
sequences $M_D^{(1)} g_D$ and $M_D^{(2)} g_D$.  Define
\begin{itemize}
\item  $\tau_F$ as the minimum over the values of $i$ for which
 $i = \iota^{(k)} (j) \; \mbox{and} \: M^{(k)}_{i,j}<1
    \; \mbox{for some} \; j \; \mbox{and} \; k$ of
$ \min \{ M^{(k)}_{i,j}-0.5, 1-M^{(k)}_{i,j} \} \cdot |g_{j+1}-g_j|$
\item  $\tau_S$ as the minimum over the values of $i$ for which
 $i = \iota^{(k)} (j) -1 \; \mbox{and} \: M^{(k)}_{i,j}>0
    \; \mbox{for some} \; j \; \mbox{and} \; k$ of
    $ \min \{ 0.5-M^{(k)}_{i,j}, M^{(k)}_{i,j} \} \cdot |g_{j+1}-g_j|$
\item  $\tau_A$ as the minimum over the values of $i$ for which
 $i = \iota^{(k)} (j) \; \mbox{and}$ \\ $ M^{(k)}_{D[i,j]}=1
    \; \mbox{for some} \; j \; \mbox{and} \; k$ of
    $|g_{j+1}-g_j|$
\item  $\tau_{A_b}$ as the minimum over the values of $i$ for which
  $i = \iota^{(k)} (j),$  \\ $M^{(k)}_{D[i,j]}=1, \; \mbox{and} \;
  M^{(k)}_{D[i+1,j+1]}>0 \; \mbox{for some} \; j \; \mbox{and} \; k$ of \\
  $0.5||g_{j+1}-g_j| - M^{(k)}_{D[i+1,j+1]}   |g_{j+2}-g_{j+1}||$
\item  $\tau_{A_e}$ as the minimum over the values of $i$ for which
  $i = \iota^{(k)} (j),$ \\ $M^{(k)}_{D[i,j]}=1, \; \mbox{and} \;
  M^{(k)}_{D[i-1,j-1]}>0 \; \mbox{for some} \; j \; \mbox{and} \; k$ of \\
  $0.5||g_{j+1}-g_j| - M^{(k)}_{D[i-1,j-1]}   |g_j-g_{j-1}||$.            
\item $\tau = \min \{ \tau_F , \; \tau_S , \; \tau_A , \; \tau_{A_b} , \; 
\tau_{A_e} \}$.
\item $Z_k \subset \{ 0, \ \dots, \ N-1 \}, \ k \in \{1, \ 2\}$, be the set
of indices for which $[M^{(k)}_D g_D]_i  = 0$ if $i \in Z_k$ and
$[M^{(k)}_D g_D]_i  \neq 0$ if $i \not\in Z_k$.
\item $B_k^F, \; k \in \{1, \ 2\}$, be the set of indices $i$ for which
  $i = \iota^{(k)} (j) \; \mbox{and} \;  M^{(k)}_{D[i+1,j]}>0$ for some $j$.
\item $B_k^S, \; k \in \{1, \ 2\}$, be the set of indices $i$ for which
  $i = \iota^{(k)} (j) \; \mbox{and}  \; M^{(k)}_{D[i-1,j]}>0$ for some $j$.
\item $B_k^{A_b}, \; k \in \{1, \ 2\}$, be the set of indices $i$ for which
  $i = \iota^{(k)} (j), \;  M^{(k)}_{D[i,j]}=1, \; \mbox{and} \; M^{(k)}_{D[i+1,j+1]}>0$ for some $j$.
\item $B_k^{A_e}, \; k \in \{1, \ 2\}$, be the set of indices $i$ for which
  $i = \iota^{(k)} (j), \;  M^{(k)}_{D[i,j]}=1, \; \mbox{and} \;
  M^{(k)}_{D[i-1,j-1]}>0$ for some $j$.
\end{itemize}
Suppose there is $v \in (0, \ \tau )$ for which
\begin{itemize}
\item $|d_k [i]| < v, \ k \in \{1, \ 2 \}$, if $i \in Z_k$,
\item $d_k [i] \geq v, \ k \in \{1, \ 2 \}$, if $i \not\in Z_k$
  and $[M^{(k)}_D g_D]_i > 0$,
  \item $d_k [i] \leq -v, \ k \in \{1, \ 2 \}$, if $i \not\in Z_k$
    and $[M^{(k)}_D g_D]_i < 0$,
  \item $|d_k [i] | > |d_k [i+1] |, \ k \in \{1, \ 2 \}$, if $i \in B^F_k$,
 \item $|d_k [i] | > |d_k [i-1] |, \ k \in \{1, \ 2 \}$, if $i \in B^S_k$.
\end{itemize}
Then the longest path scheme with edge weights
$W (i_1, \ \lambda_1 , \  i_2, \ \lambda_2 , \ v)$ identifies the location
of the first sample following each discontinuity point of $g(t)$
          within $\gamma_k$ and $y_k , \ k \in \{1, \ 2\}$.
          Moreover, it will not propose false segmentation points.

          For our motivating example, the difference sequences are
          \begin{displaymath}
\left( x,  \frac{144}{256},  \frac{112}{256}-2x,  0,  -2+2x,  0, 
 \frac{272}{256}, \frac{240}{256}-2x, 0, -2, 0,  1-2x, 0 \right)
          \end{displaymath}
          and
          \begin{displaymath}
\left(-x, 1, 0, -\frac{461}{256}+2x, -\frac{51}{256}-2x, 0, 2, 0,
- \frac{474}{256}+2x,  -\frac{38}{256},  \frac{234}{256}-2x,  \frac{22}{256},
 0 \right) .
         \end{displaymath}
          We assume that the first samples of $d_1$ and $d_2$ precede the
          samples of the support of $\tilde{g} (t)$.  Observe that
          while $x \leq \frac{102}{256}$, there is a choice of $v$ depending
          on $x$ for which the algorithm correctly aligns the sequences
          and predicts the first sample following each discontinuity point.
          Moreover, unlike cross-correlation template matching the performance degradation is less abrupt as $x$ increases.  In terms of estimating the
          amplitudes of $g(t)$ one can either work directly with the
          difference vectors or trace back to $y_1$  and $y_2$ and make
          guesses about which samples are the ones most degraded by the blur.
          The techniques of \cite{savari}-\cite{s3} can be used to make
          inferences about the locations of the discontinuity points.

\section{Conclusions} 
The accuracy of inferences made from measurements impacts the decisions made
from data.  In this work we consider how round-off errors and blur which is
Gaussian or a mixture of Gaussians can impact superresolution and image
registration when the underlying function is one dimensional, spatially
limited, and piecewise constant.  We highlight a measurement matrix and
a difference matrix in our investigations and illustrate challenges with
superresolution and the alignment technique of selecting the maximum from
cross-correlation template matching.  We find that dynamic programming can
correctly align and determine the first samples following discontinuity points
in two data sequences of noisy samples of the same signal under some conditions
on the blur, the distance between discontinuity points, and the noise.
There is a need to better understand the opportunities and limitations in
working with data coming from instruments or sensors with limited precision.

\end{document}